\documentclass[a4paper,fleqn]{cas-sc}

\usepackage[authoryear,longnamesfirst]{natbib}
\usepackage{algorithm}
\usepackage{algpseudocode}

\usepackage{longtable}

\usepackage{lipsum}
\usepackage{lineno}

\def\tsc#1{\csdef{#1}{\textsc{\lowercase{#1}}\xspace}}
\newcommand{\minus}{\scalebox{0.75}[1.0]{$-$}}
\tsc{WGM}
\tsc{QE}
\tsc{EP}
\tsc{PMS}
\tsc{BEC}
\tsc{DE}

\begin{document}

\ExplSyntaxOn
\cs_gset:Npn \__first_footerline:
  { \group_begin: \small \sffamily \__short_authors: \group_end: }
\ExplSyntaxOff 

\let\WriteBookmarks\relax
\def\floatpagepagefraction{1}
\def\textpagefraction{.001}
\shorttitle{}
\shortauthors{}

\title [mode = title]{
Exploring Machine Learning Techniques to Identify Important Factors Leading to Injury in Curve Related Crashes
}                      


\author[1]{Mehdi Moeinaddini}[
                        orcid=0000-0002-0679-3537]


\author[1]{Mozhgan Pourmoradnasseri}[orcid=0000-0002-2092-816X]
\author[1]{Amnir Hadachi}[orcid=0000-0001-9257-3858
   ]


\author[2,3,4]{Mario Cools}

\address[1]{ITS Lab, Institute of Computer Science, University of Tartu, Narva mnt 18, 51009 Tartu, Estonia}
\address[2]{LEMA Research Group, Urban \& Environmental Engineering Department, University of Li\`ege, Li\`ege, Belgium}
\address[3]{Department of Informatics, Simulation and Modelling, KULeuven Campus Brussels, Brussels, Belgium}
\address[4]{Faculty of Business Economics, Hasselt University, Diepenbeek, Belgium}



\begin{abstract}
Different factors have effects on traffic crashes and crash-related injuries. These factors include segment characteristics, crash-level characteristics, occupant level characteristics, environment characteristics, and vehicle level characteristics. There are several studies regarding these factors' effects on crash injuries. However, limited studies have examined the effects of pre-crash events on injuries, especially for curve-related crashes. The majority of previous studies for curve-related crashes focused on the impact of geometric features or street design factors. The current study tries to eliminate the aforementioned shortcomings by considering important pre-crash events related factors as selected variables and the number of vehicles with or without injury as the predicted variable. This research used CRSS data from the National Highway Traffic Safety Administration (NHTSA), which includes traffic crash-related data for different states in the USA. The relationships are explored using different machine learning algorithms like the random forest, C5.0, CHAID, Bayesian Network, Neural Network, C\&R Tree, Quest, etc. The random forest and SHAP values are used to identify the most effective variables. The C5.0 algorithm, which has the highest accuracy rate among the other algorithms, is used to develop the final model. Analysis results revealed that the extent of the damage, critical pre-crash event, pre-impact location, the trafficway description, roadway surface condition, the month of the crash, the first harmful event, number of motor vehicles, attempted avoidance maneuver, and roadway grade affect the number of vehicles with or without injury in curve-related crashes.

\end{abstract}



\begin{keywords}
Pre-Crash Events, Machine Learning, Number of Vehicles with or without Injury, Curve Related Crashes, Most Effective Variables
\end{keywords}
\maketitle

\section{Introduction}
Globally, more than 1.25 million people die per year as a result of road traffic crashes \citep{WHO2018}, and 20-50 million people suffer minor and major injuries due to motor vehicle crashes \citep{WHO2018}. Crashes involve the potential loss of human life and damage to vehicles in addition to causing extra travel costs as a result of delays in traffic \citep{alireza2002accident}. Road traffic crashes impose considerable economic and social losses on vehicle manufacturers, society, and transportation agencies \citep{haghighi2018impact}. Although during the last decade policymakers and planners have tried to reduce these losses, still more research and studies are needed to identify factors that have effects on crash injuries to reduce these social and economic losses. 

The number of injuries is twice as high on curves in comparison to straight roads \citep{chen2010mining}. Some of these curve-related crashes occurred due to drivers that cannot recognize the sharpness and presence of upcoming curves \citep{wang2017crashes}. The probability of a fatal crash at horizontal curves is significantly higher than in other segments \citep{wang2017crashes}. In 2008, around 27 percent of fatal crashes in the United States occurred at horizontal curves and most of these curve-related fatalities (over 80 percent) were in roadway departures \citep{FHWA}. So, annually more than one-quarter of all motor-vehicle fatalities in the United States are related to curve-related crashes \citep{wang2017crashes}. Because of this huge number of fatalities and injuries, the interest in curve-related crashes is significantly high. Thus, there is a need to examine the relationship between injuries and factors that have important effects on injuries in these crashes.

There are five categories of factors that have effects on traffic crash injuries. These factors include crash level factors such as crash time, crash type, cause of crash and speed \citep{hao2016effect,qin2006bayesian}, vehicle level factors such as vehicle age and type \citep{richter2005improvements, bedard2002independent, langley2000motorcycle}, occupant level factors such as the number of occupants, driver attention and alcohol involvement \citep{movig2004psychoactive, petridou2000human}, roadway design and environmental level factors such as the number of lanes, traffic control, road curvature, road grade and pavement surface \citep{moeinaddini2014relationship, moeinaddini2015analyzing, rengarasu2007effects, aarts2006driving, karlaftis2002effects, ahmed2012assessment, brijs2008studying, golob2003relationships}.

There are various studies regarding the effects of the aforementioned five categories on crash severities. \citep{duddu2018modeling} examined the effects of road characteristics, environmental conditions, and driver characteristics on driver injury severity (for both at-fault and not-at-fault drivers), using a partial proportional odds model. The results of this study show that the age of the driver, physical condition, gender, vehicle type, and the number and type of traffic rule violations have significantly higher impacts on injury severity in traffic crashes for not-at-fault drivers compared to at-fault drivers. In addition, road characteristics, weather conditions, and geometric characteristics were observed to have similar effects on injury severity for at-fault and not-at-fault drivers. Driving inattention and distracted driving behavior are two important causes of traffic crashes \citep{bakhit2019distraction}. \citep{guo2013individual} predicted high-risk drivers using personality, demographic, and driving characteristic data. The results of their study show that the driver’s age, personality, and critical incident rate have significant effects on crash and near-crash risk. Three inter-related variables, including failures of driver attention, misperceptions of speed and curvature, and poor lane positioning, are important reasons for driver errors associated with horizontal curves \citep{charlton2007role}. 

\citep{charlton2007role} used simulation to find the level of driver attention by comparing advance warning, delineation, and road marking. The results of this study show rumble strips can produce appreciable reductions in speed compared to advance warning signs. \citep{haghighi2018impact} examined the effects of different roadway geometric features (e.g. curve rate, lane width, narrow shoulder, shoulder width, and driveway density) on the severity outcomes in two-lane highways in rural areas, using data from 2007 to 2009 in Illinois. In their research, the effects of environmental conditions and geometric features on crash severity were analyzed using a multilevel ordered logit model. The results showed that the presence of a 10-ft lane and/or narrow shoulders, lower roadside hazard rate, higher driveway density, longer barrier length, and shorter barrier offset have lower severe crash risk. 

\citep{wang2017crashes} considered the effects of variables such as driver demographic and behavioral characteristics, traffic environment characteristics, and roadway design characteristics on the odds of a safety-critical event (e.g., using a cell phone, interaction with passengers, external distraction, talking or singing, reaching or moving objects, and drinking or eating) in curve related crash and near-crash events using a logistic regression model. However, some important variables, such as variables that can explain the critical events and reactions or maneuvers of drivers during the crash were missing in this study. Moreover, this study also did not consider the effects of these factors on injury severity. Some limited studies examine driver behavior on traffic safety negotiated with curve using simulators (e.g., \citep{jeong2017horizontal}; \citep{yotsutsuji2017car}; \citep{abele2011relationship}; \citep{charlton2007role}) but these studies represent just a simplified real-life situation based on some certain assumptions and validation is a challenging process in these studies.   

Various studies focused on curve characteristics and crash risk \citep{elvik2013international}. The majority of these studies investigated the effects of speed and speed limit (e.g., \citep{yotsutsuji2017car}, \citep{wang2017crashes}; \citep{dong2015assessment}; \citep{vayalamkuzhi2016influence}) and road design characteristics such as radius of the curve, curve rate and curve length on traffic safety (e.g., \citep{yotsutsuji2017car}; \citep{haghighi2018impact}; \citep{wang2017crashes}; \citep{khan2013safety}; \citep{schneider2010effects}). In addition to speed and road design factors, some researchers investigated the effects of socio-demographic factors for drivers (age, gender, income, etc.) on traffic safety in curve-related crashes (e.g., \citep{wang2017crashes}). However, only a few efforts have been undertaken to quantify the effects of pre-crash events on traffic safety and crash injuries, specifically in curve crashes \citep{wang2017crashes}. To our knowledge, little is known in the related transportation literature regarding the impacts of pre-crash events on traffic injuries for curve-related crashes \citep{bargman2017counterfactual}. The current study tries to eliminate these shortcomings by considering pre-crash events related factors as selected variables and the number of vehicles with or without injury as the predicted variable in curve-related crashes.

\section{Data and Methodology}
This research focuses on the relationship between pre-crash events related factors and the number of vehicles with or without injury in different states of the United States for curve-related crashes in 2020. In this study, the data are extracted from Crash Report Sampling System (CRSS) in the National Highway Traffic Safety Administration (NHTSA) report. This database has data for 94718 vehicles involved in crashes in different states of the United States. The data are extracted from all reliable police-reported motor vehicle traffic crashes. The database includes data for pedestrians, cyclists, and all types of motor vehicles and covers different types of crashes. In this study, vehicle-related data are used to explore the effects of pre-crash events on the number of vehicles with or without injuries for curve-related crashes. Around 8\% of the involved vehicles in these crashes are related to crashes on curves (7542 crashes out of 94718). Table 1 shows the frequency of crashes based on roadway alignment for cases with available injury levels (90269 crashes out of 94718). This table indicates the proportion of involved vehicles in curve-related crashes for vehicles without or with injuries. 

\begin{table}[width=.9\linewidth,cols=4,pos=h]
\caption{Frequency of crashes based on roadway alignment.}\label{tbl1}
\begin{tabular*}{\tblwidth}{@{} CCCCCCCCCC@{} }
\toprule
&\textbf{Injury}\\
\midrule
&	No	& & & &			Yes	& & & &			Total\\
\midrule
\multirow{2}{*}{\textbf{Roadway Alignment}}  & &Row\%&	Col\%&	Cell\%&	&	Row\%&	Col\%&	Cell\%&	No.\\
\cline{2-10}
 ~ &1,864&	81&	3&	2&	428&	19&	1&	0&	2,292 \\
\midrule
Non-Trafficway &	51,97&	67&	87&	58&	25,604&	33&	84&	28&	77,574\\
\midrule
Straight&	1,758&	55&	3&	2&	1,436&	45&	5&	2&	3,194\\
\midrule
Curve Right	&1,502&	49&	3&	2&	1,581&	51&	5&	2&	3,083\\
\midrule
Curve Left&	563	&57&	1&	1&	429&	43&	1&	0&	992\\
\midrule
Curve-Unknown Direction&	1,998&	65&	3&	2&	1,085&	35&	4&	1&	3,083\\
\midrule
Not Reported&	35&	69&	0&	0&	16&	31&	0&	0&	51\\
\midrule
Total & 59,69 &	66 & 100&	66&	30,579&	34&	100	&34&	90,269 \\
\bottomrule
\end{tabular*}
\end{table}

Different pre-crash variables are considered predictors for the predicted variable, i.e., the number of vehicles with or without injury (0: vehicle without injury, 1: vehicle with injury) in the crashes that occurred in a curve. Table 2 defines these variables after decoding. The selected variables in Table 2 present pre-crash events in addition to the driver behavior and some crash level data, such as the month of the crash. Since the crash rate may differ in different seasons, the month of the crash indicates the crashes in the winter. In addition, since more occupants in crashes may lead increased chance of vehicle injury, the effect of the number of occupants in each vehicle is also tested in this study. 

Some driver behavior-related factors may affect the number of vehicles with or without injury. These factors include driver errors (e.g., careless driving, aggressive driving, improper or erratic lane changing and overcorrecting), driving too fast, and avoidance maneuvers. Some factors represent environment and vehicle conditions. For example, the vehicle’s manufacturing year represents the vehicle’s condition assuming that newer and less damaged cars may lead to fewer injuries because of more advanced safety facilities. The driving environment condition is presented by traffic way description (one way or two ways and how the traffic ways are separated), speed limit, roadway surface condition, and the presence of traffic controls.

The rest of the selected variables, such as the harmful events, the critical pre-crash events, the vehicle's stability after the critical event, the location of the vehicle after the critical event, and the crash type, represent pre-crash events. As expected, all required data for the main variables are not reported for all curve-related crashes in the NHTSA report, and there are many not reported or unknown data for these variables. In addition, to focus on curve-related crashes, only the vehicles that negotiate a curve prior to realizing an impending critical event or just prior to impact are included. Therefore, after data preparation (removing incomplete information for the main variables and including the vehicles that negotiate a curve), the total number of curve-related crashes retained in this study equals 740. 


\begin{longtable}{ p{.20\textwidth} p{.20\textwidth} p{.25\textwidth}} 
\caption{Selected variables.} 
\label{tab:IDv}\\
\toprule
\textbf{Variable}&	\textbf{Description}& 	\textbf{Coding}\\
\midrule
Urban or rural&	\multicolumn{1}{p{6cm}}{The geographical area of the crash is essentially urban or rural}&	\makecell{\multicolumn{1}{p{4cm}}{1: urban}\\\multicolumn{1}{p{4cm}}{2: rural} }\\
\midrule
\makecell{Number of motor\\ vehicles}&	\multicolumn{1}{p{6cm}}{Number of motor vehicles involved in the crash}&	\makecell{\multicolumn{1}{p{4cm}}{0: 1}\\\multicolumn{1}{p{4cm}}{1: >1}}\\
\midrule
\makecell{Number of \\occupants}&	\multicolumn{1}{p{6cm}}{The number of occupants in each vehicle}&	\makecell{\multicolumn{1}{p{4cm}}{0: 1}\\\multicolumn{1}{p{4cm}}{1: >1 }}\\
\midrule

\makecell{First harmful\\ event}&	\multicolumn{1}{p{6cm}}{The first injury or damage producing event}&	\makecell{\multicolumn{1}{p{4cm}}{0: other events}\\\multicolumn{1}{p{4cm}}{1: collision with motor vehicles in transport}  }\\
\midrule
Vehicle’s Model&	\multicolumn{1}{p{6cm}}{Manufacturer's model year of the vehicle}&	\makecell{\multicolumn{1}{p{4cm}}{0: <2010}\\\multicolumn{1}{p{4cm}}{1: >=2010}} \\
\midrule
\makecell{Initial contact\\ point}&	\multicolumn{1}{p{6cm}}{The area on the vehicle that produced the first instance of injury or damage}& \makecell{	\multicolumn{1}{p{4cm}}{0: other areas}\\\multicolumn{1}{p{4cm}}{1: front}} \\
\midrule
\makecell{Extent of \\damage}&	\multicolumn{1}{p{6cm}}{The amount of damage sustained by the vehicle}&	\makecell{\multicolumn{1}{p{4cm}}{0: Not disabling damage}\\\multicolumn{1}{p{4cm}}{1: Disabling damage }}\\
\midrule
\makecell{Most harmful\\ event}&	\multicolumn{1}{p{6cm}}{The event that resulted in the most severe injury or the greatest damage}&	\makecell{\multicolumn{1}{p{4cm}}{0: other events}\\\multicolumn{1}{p{4cm}}{1: collision with motor vehicles in transport} } \\
\midrule
Speeding-related&	\multicolumn{1}{p{6cm}}{The driver’s speed was related to the crash}& \makecell{	\multicolumn{1}{p{4cm}}{0: no}\\\multicolumn{1}{p{4cm}}{1: yes}} \\
\midrule
Driver Error&	\multicolumn{1}{p{6cm}}{Factors related to the driver errors expressed by the investigating officer}&	\makecell{\multicolumn{1}{p{4cm}}{0: no error}\\ \multicolumn{1}{p{4cm}}{1: error (e.g., careless driving or aggressive driving/road rage or operating the vehicle in an erratic, reckless, or negligent manner, improper or erratic lane changing or improper lane usage, or driving on the wrong side of two-way traffic way, etc.)} }\\
\midrule
Traffic way&	\multicolumn{1}{p{6cm}}{Trafficway description}	&\makecell{\multicolumn{1}{p{4cm}}{0: divided two-way and others}\\
\multicolumn{1}{p{4cm}}{1: not divided two-way}}  \\
\midrule
Speed limit	&\multicolumn{1}{p{6cm}}{The posted speed limit in miles per hour}	&\makecell{\multicolumn{1}{p{4cm}}{0: <46}\\
\multicolumn{1}{p{4cm}}{1: >=46} }\\
\midrule
Roadway alignment&	\multicolumn{1}{p{6cm}}{The roadway alignment prior to the critical pre-crash event}&	\makecell{\multicolumn{1}{p{4cm}}{1: curve right}\\
\multicolumn{1}{p{4cm}}{2: curve left} \\
\multicolumn{1}{p{4cm}}{3: curve – unknown direction }}\\
\midrule
Grade&	\multicolumn{1}{p{6cm}}{Roadway grade prior to the critical pre-crash event}&	\makecell{\multicolumn{1}{p{4cm}}{0: not level}\\
\multicolumn{1}{p{4cm}}{1: level}}
\\
\midrule
Roadway surface condition&	\multicolumn{1}{p{6cm}}{Roadway surface condition prior to the critical pre-crash event	}&\makecell{\multicolumn{1}{p{4cm}}{0: not dry}\\
\multicolumn{1}{p{4cm}}{1: dry}}\\
\midrule
Traffic control device&	\multicolumn{1}{p{6cm}}{The presence of traffic controls in the environment prior to the critical pre-crash event}&	\makecell{\multicolumn{1}{p{4cm}}{0: no}\\
\multicolumn{1}{p{4cm}}{1: yes}}\\
\midrule
Critical pre-crash event &	\multicolumn{1}{p{6cm}}{The critical event which made this crash imminent }&\makecell{	\multicolumn{1}{p{4cm}}{1: the vehicle itself (loss of control, traveling too fast, etc.)}\\
\multicolumn{1}{p{4cm}}{2: other vehicles (traveling in the opposite direction, encroaching into the lane, etc.)}\\
\multicolumn{1}{p{4cm}}{3: others (pedestrian in the road, animal approaching road, etc.)}}\\
\midrule
Attempted avoidance maneuve r&\multicolumn{1}{p{6cm}}{Movements/actions taken by the driver within the crash}&	\makecell{\multicolumn{1}{p{4cm}}{0: no action}\\
\multicolumn{1}{p{4cm}}{1: braking}\\
\multicolumn{1}{p{4cm}}{2: others.}} \\
\midrule
Pre-impact stability&	\multicolumn{1}{p{6cm}}{The stability of the vehicle after the critical event but before the impact.}&	\makecell{\multicolumn{1}{p{4cm}}{0: no tracking (skidding, loss-of-control, etc.)}\\
\multicolumn{1}{p{4cm}}{1: tracking}} \\
\midrule
Pre-impact location	&\multicolumn{1}{p{6cm}}{The location of the vehicle after the critical event but before the impact}&\makecell{	\multicolumn{1}{p{4cm}}{0: not departed roadway}\\
\multicolumn{1}{p{4cm}}{1: departed roadway}}
\\
\midrule
Crash Type&	\multicolumn{1}{p{6cm}}{The type of crash }&	\makecell{\multicolumn{1}{p{4cm}}{0: others}\\
\multicolumn{1}{p{4cm}}{1: single driver involved (roadside departure, collision with pedestrians, etc.)} }\\
\bottomrule
\end{longtable}

Different methods like multinomial logit models, general linear models, order prohibit models, linear regression (\citep{clark2004rural}; \citep{levine1995spatial}; \citep{abdel2003analysis}; \citep{yang2011exploring}; \citep{fan2015analyzing}), Negative binomial (NB) models (\citep{abdel2000modeling}; \citep{hadayeghi2003macrolevel}; \citep{hadayeghi2007safety}; \citep{wei2013empirical}; \citep{moeinaddini2014relationship}; \citep{moeinaddini2015analyzing}), Poisson models (\citep{movig2004psychoactive}) and Zero-inflated Poisson and NB models (\citep{qin2004selecting}; \citep{shankar1997modeling}) have been used to analyze traffic fatalities and injuries related data. In addition to these methods, some studies have used decision tree approaches (e.g., ID3, C4.5, C5.0, C\&R, CHAID) to find the major contributing factors to collision and the number of fatalities. For example, the Classification and Regression (C\&R) Tree was used by \citep{tavakoli2011data}. One of the most common approaches for representing classifiers is Decision Trees \citep{maimon2005data}. Researchers from different disciplines such as machine learning, statistics, pattern recognition, and data mining use decision trees to analyze data in a more comprehensive way \citep{maimon2005data}.  

\citep{zhang2013decision} used data mining models using ID3 and C4.5 decision tree algorithms to evaluate the traffic collision data in Canada. \citep{chong2005traffic} compared different machine learning paradigms, including neural networks trained using hybrid learning approaches, support vector machines, decision trees, and a concurrent hybrid model involving decision trees and neural networks to model the injury severity of traffic crashes. The results of their study show that for the non-incapacitating injury, the incapacitating injury, and the fatal injury classes, the hybrid approach performed better than a neural network, decision trees, and support vector machines. \citep{da2017identification} used the C\&R algorithm as a useful tool for identifying potential sites of crashes with victims. \citep{chang2006analysis} used C\&R to find the relationships between crash severity with factors such as drivers’ and vehicles’ variables and the road and environment characteristics. The results of their study show that vehicle type is one of the most important factors that have an effect on the severity of the crash.

The majority of traditional and parametric analysis techniques have different assumptions and pre-defined functions that describe the relationship between the selected and the predicted variables \citep{chang2006analysis}. If these assumptions are violated, the model power can be affected negatively \citep{griselda2012using}. Therefore, assumption-free models such as decision trees can be used to avoid this limitation \citep{griselda2012using}. \citep{kuhnert2000combining} compared the results of different methods such as logistic regression, multivariate adaptive regression splines (MARS), and C\&R in the analysis of data related to the injury in motor vehicle crashes. The findings of their study show the usefulness of non-parametric techniques such as C\&R and MARS to provide more attractive and informative models \citep{griselda2012using}. \citep{pandya2015c5} compared the results of ID3, C4.5, and C5.0 with each other. They found that among all these classifiers, C5.0 gives more efficient, accurate, and fast results with low memory usage (fewer rules compare to other techniques). 

Finding the most accurate prediction models can help planners and designers to develop better traffic safety control policies. In the current study, a variety of modeling techniques is envisaged as possible analysis methods, and the most appropriate model based on the accuracy rate is retained for further discussion of the results. 

The first step to finding the most appropriate model is applying random forest to identify the most influential variables among the selected variables. Then, the identified effective variables are used as selected variables to explore the effects of these selected variables on the predicted variable.  Random forest is a common method for selecting the most effective variables in studies with a high number of predictors (\citep{jahangiri2016red}; \citep{kitali2018likelihood}; \citep{zhu2018design}; \citep{aghaabbasi2020predicting}; \citep{lu2020hybrid}). The random forest aggregates many binary decision trees. 

Cross-validation (10-fold cross-validation) which generally results in a less biased model than other methods like train and test split is applied to estimate the accuracy. Random and grid search for hyper-parameter optimization \citep{bergstra2012random} are used for hyper-parameter optimization. After applying random forests, the SHAP (SHapley Additive exPlanations) values \citep{lundberg2017unified} are used to select the most important variables. The SHAP explains the contribution of each observation and provides local interpretability but the traditional importance values explain each predictor’s effects that are based on the entire population. After estimating SHAP values, the not-important variables are excluded one by one. The accuracy rates for each step of exclusion are used to find the most effective variables.

\subsection{Models Description}
 A couple of machine-learning algorithms were explored to identify the relationships between curve-related crashes and pre-crash events. To this end, identifying the significant features for training the models is an important step to ensure a good training process and better results.
 
\subsubsection{Feature selection}
Approximating the functional relationship between the input data and the output is one of the fundamental problems when applying machine learning methods. Selecting the significant feature for training the machine learning models is crucial to avoid overfitting and to induce high computational costs. Therefore, our approach utilized the power of random forest as a classifier and interpreted with Shapely values. Relying on SHAP values helps to perform feature selection based on ranking. This means that instead of using the embedded feature selection process of the random forest, we use the SHAP value to select the ones with the highest shapely values. This approach's advantage is avoiding any bias in the native tree-based feature built by the random forest approach. 

\subsubsection{C5.0 decision tree algorithm}
Decision trees are built using recursive partitioning. The algorithm starts by creating the root node, which in our case, is the actual data. Then, based on the most significant feature selected, the data is partitioned into groups. These groups are the distinct value of this feature, and this decision forms the first set that constitutes the tree branches. The algorithm divides the nodes until the criterion is reached (Algorithm \ref{alg-c5}). In practice, the C5.0 algorithm decides the split by using the concept of entropy for measuring purity. This means for a segment of data $S$, if the entropy is close to value $0$ indicates that the data sample is homogenous, and the opposite if it is close to value $1$ and it is defined as follows:  

\begin{equation}
    \text{Entropy}(S)=\sum_{i=1}^{m} \minus P_i\log_2 P_i;
\label{eq-entropy}
\end{equation}

where $m$ refers to the number of different class levels, and $p_i$ is the proportion of values falling into the class level $i$. However, even after conducting this step, to understand the homogeneity of the data, the algorithm still needs to decide how to split the set. To solve this, the C5.0 uses the entropy to spot the variations of homogeneity resulting from the split. This measure is called \textit{Information Gain}, which is defined using equation \ref{eq-gain}. It quantifies the gained information of an attribute $A$, when selecting data set $S$ .

\begin{equation}
    \text{InfoGain}(S,A)= \text{Entropy}(S)~ \minus \sum_{v\in V(A)} \frac{|S_v|}{|S|}\text{ Entropy}(S_v);
\label{eq-gain}
\end{equation}

 where $V(A)$ is the set of all possible attribute values for $A$, and $S_v$ is the subset of $S$ for which $A$ has value $v$. Hence, the creation of homogeneous groups after the split on specific feature is better when the information gain is high. 

\begin{algorithm}
\caption{C5.0 decision tree }\label{alg:cap}
\label{alg-c5}
\begin{algorithmic}
\Require $ \text{Data} \ \ T=\{(x_i,y_j), \ \ i,j\in \{1,2,\cdots,n\}\}$ 
\Require $\text{Attributes} \ \ A=\{a_l, \ \ l\in \{1,2,\cdots,p\}\}$ 
\Require $\text{InfoGain}=\{g_k, \ \ k\in \{1,2,\cdots,d\}\}$ 
\State Create node  $N$
\If{$S$ are all the same class, $C$ }  
    \State label  $N$  with class $C \rightarrow N^{(C)}$\\
    \Return $N^{(C)}$ as leaf node
\EndIf
\If{$A \neq \emptyset$  or  $a_i=a_j$ for all $a_i,a_j \in A$}
    \State label $N$ with  majority class $M$ in $S \rightarrow  N^{(M)}$
    \Return $N^{(M)}$  as leaf node 
\EndIf
\State select best attribute $a_i$ using InfoGain
\For {every  $a_i^v \in  a_i$}
    \State label node  $N$ with splitting criterion
    \If{$S_v \neq \emptyset$} where $S_v$ is the set of data  in $S$ equal to $a_i^v$ 
         \State label  $N$ with majority  class  $M$ in  $S$ $\rightarrow  N^{(M)}$
         \Return $N^{(M)}$ as  leaf node 
    \Else \ \
         \Return $N$ with splitting criterion $(S_v, A\{a_i\})$
    \EndIf 
\EndFor
\end{algorithmic}
\end{algorithm}

\subsubsection{Chi-squared Automatic Interaction Detection}
Chi-squared automatic interaction detection (CHAID) is one of the techniques based on a tree machine learning algorithm. Hence, the algorithm relies on the multiway split using Chi-square or F-test. The CHAID algorithm uses two approaches for separation reference depending on the variable. If the variable is categorical, Pearson’s Chi-square is adequate; otherwise, the likelihood ratio Chi-square statistic. The CHAID uses Pearson's Chi-squared test of independence to test the existence of an association between two categorical variables (``true" or``false").
The main steps to calculate Chi-square for the split are as follows:
\begin{enumerate}
    \item Calculating the deviation for "true" and "false" in the node which constitutes the Chi-square computation.
    \item Getting the split by computing the sum of all the chi-square of "true" and "false" of each split node.
\end{enumerate}



\subsubsection{Classification and Regression Tree node} 
The Classification and Regression (C\&R) Tree node algorithm is a classification algorithm that is based on a binary tree built by splitting nodes into two child nodes continually similarly to C5.0 method. The algorithm is designed in such a manner that it follows three major steps: 
\begin{enumerate}
    \item Identifying each feature's best split.
    \item Identifying the node's best split.
    \item Based on the step 2 result, the node is split and repeats the process from step 1 till the stopping criterion is met.  
\end{enumerate}
For performing the split, Gini's impurity index criterion is used and it is defined as follows for a node $t$:
\begin{equation}
    \text{Gini }(t)=\sum_{i,j} C(i \mid j) P(i\mid n) P(j\mid n);
\end{equation}
where, 
\begin{itemize}
    \item     $C(i\mid j)$ is the cost of classifying wrongly a class $j$ as a class $i$ and it is defined as follows:
\begin{equation}
  C(i \mid j) =
  \begin{cases}
    1 & \text{$i \neq j$} \\
    0 & \text{$i = j$}
  \end{cases}
\end{equation}
\item $P (i\mid n)$ is the probability of $i$ falls into node $n$
\item $P (j\mid n)$ is the probability of $j$ falls into node $n$
\end{itemize}
The splitting criterion is based on Gini's impurity criterion, which follows a decrease of impurity using the following formula: 
\begin{equation}
    \Delta \text{Gini }(s,n)= \text{Gini }(t) \minus P_R \text{Gini }(n_R) \minus P_L \text{Gini }(n_L);
\end{equation}
where, $\Delta \text{Gini }(s,n)$ is the decrease of impurity at node $n$ with a split $s$. $P_R$ and $P_L$ are, respectively, the probabilities of sending the case to the right or the left node $n_R$ or $n_L$, and $\text{Gini }(n_R)$ and $\text{Gini }(n_L)$ are respectively the Gini impurity index of the right and left child node.

\subsubsection{Bayesian network }
A Bayesian network is a compact graphical interpretation of the causal relationship between variables of a dataset. The structure is presented by a directed acyclic graph (DAG), and parameters are expressed as conditional probabilities. In order to learn the network, structure and conditional probabilities must be known. The structure is learned by DAG search algorithms and assigning prior probabilities. Then parameters are determined by the maximum likelihood estimation. Including the prior knowledge of the causal structure of DAG is a crucial step in learning parameters in this method. 

\subsubsection{Logistic regression}
Logistic regression is a statistical classification algorithm that maps the results of a linear function onto the regression function 
\begin{equation}
    P({\bf X})=\frac{e^{\beta_0 + \bf{\beta X}}}{1+e^{\beta_0+\bf{\beta X}}}.
\end{equation}
Based on the maximum likelihood method, the coefficients $\beta_0$ and $\beta$ are estimated in the training phase. 
This algorithm is a suitable classifier when variables can be categorized into two or a few classes. In contrast to linear regression, the logistic function associates probabilities to each possible output class by producing an S-shape curve and a range of output between 0 and 1.

\subsubsection{Neural Network} 
Our case study adopted neural network architecture based on five hidden layers using a multilayer perception model. The multilayer perceptron is the most straightforward feed-forward network. When the layers are increased, it can provide exciting performance in learning and precision. The units are arranged into a set of layers, and each layer contains a collection. The first layer is the input layer, populated by the value of input features. Then, the input is later connected to a very hidden layer in a fully connected fashion. The last layer is the output layer, which has one unit for each network output weight with a stopping rule on the error generated.

\subsubsection{QUEST algorithm} 
Quick Unbiased Efficient Statistical Tree (QUEST) is a cost-effective classification method for building binary decision trees for categorical and quantitative predictors with a large number of variables. Instead of examining all possible splits, QUEST uses statistical analysis and a multi-way chi-square test to select the variable at each node. This leads to a significant reduction in the time complexity, compared to methods like R\&C Tree, by avoiding inefficient splits. Moreover, the split point at each node is selected based on a quadratic discriminant analysis of potential categories.   

\subsubsection{Decision List}
Decision lists are a representation of Boolean functions that work as a collection of rule-based classifiers. Rules are learned sequentially and based on a greedy approach by identifying the rule that covers the maximum number of instances in the input space $X$. Then rules are appended to the decision tree one at a time, and the corresponding data is removed from the data set in each process. A new instance is classified by examining the rules in order, and if no rule is satisfied, the default rule is applied.  

Features are defined as Boolean functions $f_i$ that map the input space $X$ onto $\{0,1\}$. For a given set of features $\mathcal{F}=\{f_i(x)\}$, with $x\in X$, and the training set $\mathcal{T}$, learning algorithm returns a selection of features $\mathcal{F'}\subset \mathcal{F}$. Once the effective features are determined, for an arbitrary input $x$, the output of the decision tree is calculated according to a set of conditions to be satisfied.



\section{Results}
The overall accuracy for the applied random forest model with all predictors is 0.67. However, the overall accuracy for the applied random forest model with the 10 most important predictors based on SHAP values can reach 0.68. Therefore, these 10 predictors are selected as the most effective variables among the selected variables (the extent of the damage, critical pre-crash event, pre-impact location, the trafficway description, roadway surface condition, the month of the crash, the first harmful event, number of motor vehicles, attempted avoidance maneuver, and roadway grade). The identified effective variables are used as selected variables for a variety of possible modeling methods to find the most appropriate model based on the accuracy rate. The accuracy of the traditional logistic regression model is lower than non-parametric models like C5.0, CHAID, C\&R Tree, and Bayesian network. In addition, potential high correlations between some of the selected crash-related variables in this study may lead to a multicollinearity concern. Therefore, it is better to use modeling techniques that can handle multicollinearity issues to be able to consider the effects of these variables. Since non-parametric models can handle multi-collinearity issues in crash-related data better than traditional and parametric models, based on the overall accuracy that is achieved for each model (refer Table 3), the C5.0 model with the highest accuracy score is used for modeling the most important predictors. To develop this C5.0 model, the minimum number of records per child branch number is considered to be 2 and the pruning severity is considered to be 75. To collapse weak subtrees, trees are pruned in local and global pruning stages. Cross-validate is used to estimate the accuracy of the model. This technique uses a set of models using subsets of the data to estimate the accuracy. C5.0 is an improved version of C4.5 that is an extension of ID3 algorithm (\citep{quinlan19934}; \citep{witten2011practical}; \citep{kotsiantis2007supervised}; \citep{quinlan1996improved}).

\begin{table}[width=.9\linewidth,cols=4,pos=h]
\caption{The overall accuracy of the possible analysis methods.}\label{tbl1}
\begin{tabular*}{\tblwidth}{@{} LL@{} }
\toprule
\textbf{Applied model}&	\textbf{Overall accuracy (\%)}\\
\midrule
C5.0 &	71.757 \\

CHAID &	70.135 \\

C\&R Tree & 68.784 \\

Bayesian Network&	67.973 \\

Logistic Regression &	66.486 \\

Neural Network	&65.27 \\

Quest &	63.514 \\

Decision List &	63.108 \\
\bottomrule
\end{tabular*}
\end{table}

The applied C5.0 model is shown in Figure1. This figure shows the total percentage and the classification of the predicted variable for each node. The overall accuracy based on the results is more than 71\%. Sixteen terminal nodes (the bottom nodes of the decision tree) have been shown in Figure 1 and it is clear that this model has 8 splitters, i.e. the extent of the damage, first harmful event, the month of the crash, critical pre-crash event, pre-impact location, roadway surface condition, the trafficway description, and roadway grade. The most important variable for data segmentation is the extent of the damage. The probability of having vehicles without injury in curve-related crashes is high in node 1. Node 1 shows that not disabling damage results in a higher rate for vehicles without injury. In contrast, from node 23 one can depict that disabling damage results in a higher rate for vehicles with injury. 
The findings show that all environmental and pre-crash events that lead to driving with extra caution are related to vehicles with a lower chance of having injuries in curve-related crashes. For example, for vehicles that have disabling damage (refer node 23), the model prediction is with injury if the month of the crash is not in the winter. The same prediction can be expected for the months in the winter if the surface is dry (refer node 27) and the pre-impact location is departed the roadway (refer node 29). However, the prediction for months in the winter can be without injury if the surface is not dry (refer node 26) or the pre-impact location has not departed the roadway (refer node 28) for dry surfaces. 
The effects of driving with extra caution can also be noticed for vehicles that do not have collisions with motor vehicles in transport. Node 1 is divided into node 2 and node 20 which are related to the first harmful event. For vehicles that have no disabling damage (refer node 1) and do not have collisions with motor vehicles in transport (refer node 2), the model prediction is without injury if the critical pre-crash event is related to the vehicle itself (refer node 3), the surface is not dry (refer node 4), and the month of the crash is in the winter (refer node 8). The same prediction can be expected for the months that are not in the winter if the roadway is not a divided two-way road (refer node 7).  The same prediction also can be expected for the critical pre-crash event that is related to the other vehicles (node 10) while the pre-impact location is departed the roadway (refer node 12) and for the other critical pre-crash events while the pre-impact location is not departed the roadway (refer node 14). The prediction is also without injury for departed the roadway in this case (refer node 15) if the roadway is a divided two-way road (refer node 16) or a not level road (refer node 18) for a divided two-way road (refer node 17).
For vehicles that have no disabling damage (refer node 1) and do not have collisions with motor vehicles in transport (refer node 2), the model prediction is with injury if the critical pre-crash event is related to the vehicle itself (refer node 3) and the surface is dry (refer node 9). The same prediction can be expected for the critical pre-crash event that is related to the other vehicles (node 10) while the pre-impact location is not departed the roadway (refer node 11). 
For vehicles that have no disabling damage (refer node 1) and have collisions with motor vehicles in transport (refer node 20), the model prediction is without injury if the pre-impact location is not departed the roadway (refer node 21) and with injury if the pre-impact location is departed the roadway (refer node 22). This finding shows that departing the roadway is a very important factor for collisions with motor vehicles in transport in curve-related crashes.
The results confirm that out of all input selected variables, eight main variables play an important role in vehicles with or without injury in curve-related crashes. Table 4 shows the importance of these main predictors based on the proposed C5.0 algorithm. Higher importance scores mean a greater contribution of the variable in predicting the number of vehicles with or without injury. A breakdown of prediction accuracy is also estimated (refer Table 5). 

\begin{figure}[!tbh]
	\centering
		\includegraphics[scale=.35]{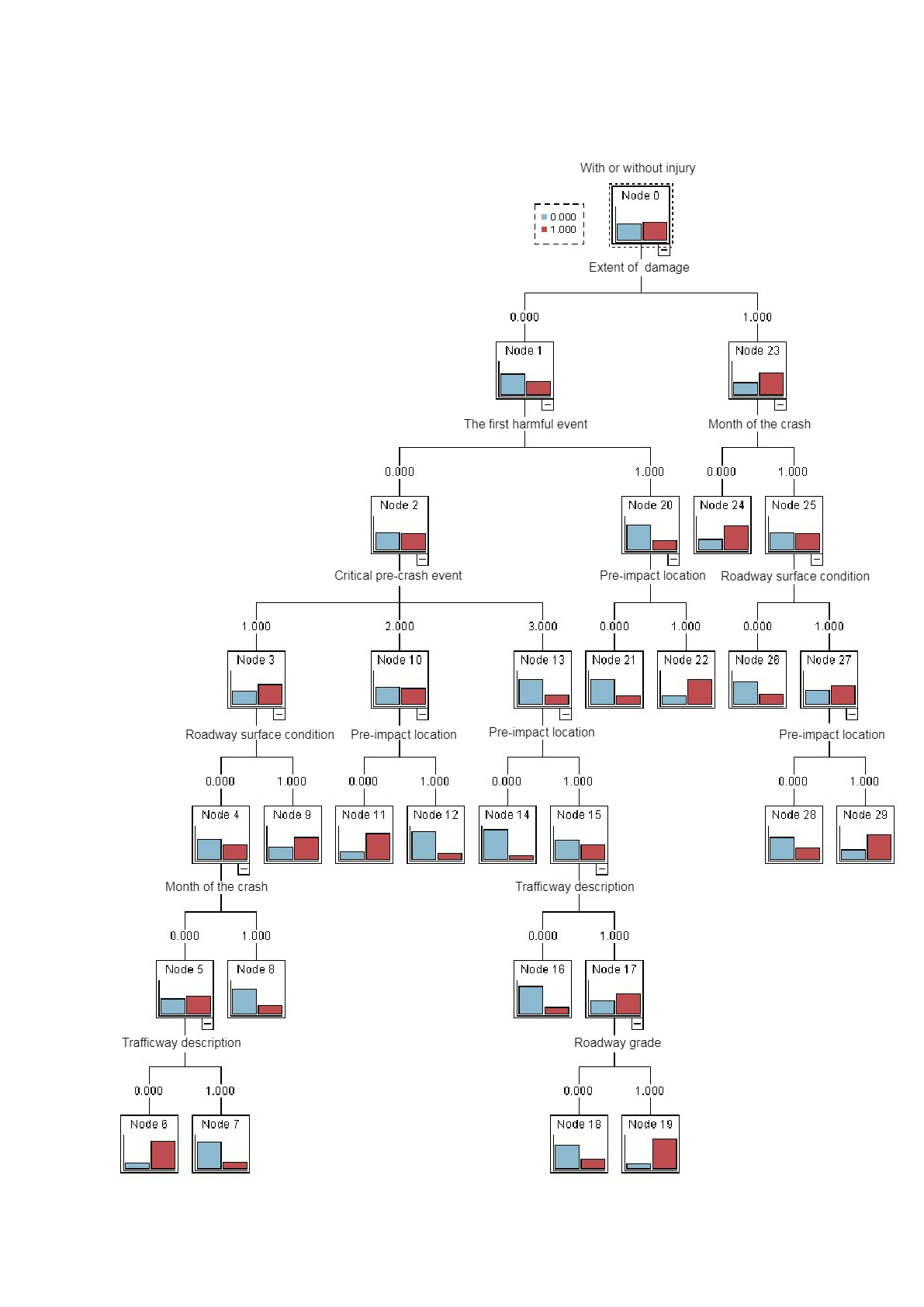}
	\caption{The proposed C5.0 model.}
	\label{FIG:1}
\end{figure}

\begin{table}[width=.9\linewidth,cols=4,pos=h]
\caption{Importance of the predictors based on the proposed C5.0 algorithm.}\label{tbl1}
\begin{tabular*}{\tblwidth}{@{} LL@{} }
\toprule
\textbf{Nodes}	&\textbf{Importance} \\
\midrule
Extent of damage&	0.3401 \\
Pre-impact location&	0.2303 \\
The first harmful event&	0.2056 \\
Month of crash&	0.1021\\
Roadway surface condition&	0.0433 \\
Trafficway description&	0.0430 \\
Roadway grade&	0.0351\\
Critical pre-crash event&	0.0006\\
\bottomrule
\end{tabular*}
\end{table}

\begin{table}[width=.9\linewidth,cols=4,pos=h]
\caption{Coincidence matrix for predicted values .}\label{tbl1}
\begin{tabular*}{\tblwidth}{@{} LLLL@{} }
\toprule
&\textbf{0}	&\textbf{1}&\textbf{	\%}\\
\midrule
\textbf{0}	&206&	140&	60 \\
\textbf{1}	&69	&325&	82 \\
\bottomrule
\end{tabular*}
\end{table}

\section{Discussion and Conclusion}
To reduce the number of crash injuries and have better planning decisions and strategies, it is important to have deep knowledge about factors influencing crash injuries. The proposed C5.0 algorithm (with a higher overall accuracy rate compared to the other analysis methods) can help to identify the variables that have the most important impacts on the number of vehicles with or without injury in curve-related crashes. This study used the 2020 NHTSA data for different states in the USA to find the key variables that affect the number of vehicles with or without injury in curve-related crashes. The results show that the extent of the damage, critical pre-crash event, pre-impact location, the trafficway description, roadway surface condition, the month of the crash, the first harmful event, number of motor vehicles, attempted avoidance maneuver, and roadway grade affect the number of vehicles with or without injury the most. 
The C5.0 model shows that most of the important predictors are related to environmental and pre-crash events that lead to driving with extra caution. Analysis results also revealed that departing the roadway is a very important factor for collisions with motor vehicles in transport in curve-related crashes. This is in line with previous studies like \citep{wang2017crashes} that identified traveling too fast on curves as one of the most important factors that contribute to crash fatalities. \citep{wang2017crashes} considered the effects of driver behavior factors such as speeding in curve-related crashes. Still, this study did not consider factors such as critical events and pre-critical event factors in addition to the reaction or maneuvers of the driver during the crash.

In \citep{wang2017crashes}, the icy and snowy road surface is another important factor that was associated with curve-related crashes, however, in our study, not dry surface was a significant factor for vehicles with injury for the months that are not in the winter. This is in line with the \citep{eisenberg2005effects} study that evaluated the impacts of snowy surfaces on traffic crash rates in the USA (1975-2000). They found that snow days are associated with fewer severe crashes, whereas more no severe crashes and property-damage crashes are reported on snow days. Therefore, although icy and snowy surfaces can be an important factor in the crash rate, they do not have a high association with severe crashes. 
Crash type is another important variable in similar research such as \citep{da2017identification} and \citep{griselda2012using}. However, the proposed models in the current study show that crash type is not significant while considering critical pre-crash events. Based on the proposed final model, the extent of damage, the pre-impact location, the first harmful event, and the critical pre-crash event are among the significant pre-crash events that can affect the number of vehicles with or without injury in addition to the environmental factors like the month of the crash, roadway surface condition, the traffic way description, and roadway grade. There are limited studies about the impacts of these important pre-crash events and environmental factors on traffic injuries \citep{bargman2017counterfactual} and although curve related crashes are associated with a high proportion of severe crashes, there is no study about the effects of these important factors on the number of vehicles with or without injury in curve-related crashes. 
Applying non-parametric tree-based models like C5.0 has some advantages compared to traditional regression and other parametric models. \citep{chang2006analysis} highlighted that C5.0 analysis does not require the specification of a functional form and also it can handle multi-collinearity problems, which often occur due to the high correlations between selected variables in traffic injury data (e.g., collision type and driver/vehicle action; weather condition and pavement condition). The proposed C5.0 model can be presented graphically, which is intuitively easy to interpret without complicated statistics. It also provides useful results by focusing on limited, yet most influential factors \citep{chang2006analysis}.  
However, non-parametric models have some disadvantages such as a lack of formal statistical inference procedures \citep{chang2006analysis}. These models also do not have a confidence interval for the risk factors (splitters) and predictions \citep{chang2006analysis}. The structure and accuracy can be changed significantly if different partitioning and sampling strategies (e.g., stratified random sampling) are applied for model testing. It is not recommended to have a generalization based on the results of nonparametric techniques. Therefore, the tree models are often applied to identify important variables, and other modeling techniques are needed to develop final models. Since sampling and different partitioning strategies are not applied to the proposed models in this study, this disadvantage is not a great concern for the current research. 


\section*{Acknowledgments}
This work was supported by the European Social Fund via IT Academy programme, and the Estonian Centre of Excellence in IT (EXCITE).

\bibliographystyle{cas-model2-names}

\bibliography{main}





\end{document}